\documentclass[acmtog]{acm}
\usepackage{color}
\usepackage{subfigure}
\usepackage{units}
\usepackage{expl3}
\usepackage{amsmath}
\usepackage{multirow}
\usepackage{colortbl}
\usepackage{algorithm}
\usepackage[noend]{algpseudocode}

\definecolor{Yellow}{rgb}{1,1, 0.6}
\definecolor{Red}{rgb}{1, 0.6, 0.6}
 
\usepackage{booktabs}
\setcitestyle{square}

\renewcommand\footnotetextcopyrightpermission[1]{}
\makeatletter
\renewcommand\@formatdoi[1]{\ignorespaces}
\makeatother

\begin{document}

\title{Cycle-consistent Generative Adversarial Networks for Neural Style Transfer using data from Chang'E-4}

\author{J. {de Curt\'o i D\'iAz} and R. A. Duvall.}
\affiliation{\institution{\\Iris Lunar Rover. Carnegie Mellon.}}

\renewcommand{\shortauthors}{De Curt\'o i D\'iAz and Duvall.}
\renewcommand{\shorttitle}{Cycle-consistent Generative Adversarial Networks for Neural Style Transfer using data from Chang'E-4}

\authorsaddresses{decurtoidiaz@ieee.org, rduvall@andrew.cmu.edu}

\begin{teaserfigure}
\centering
\includegraphics[width=\textwidth]{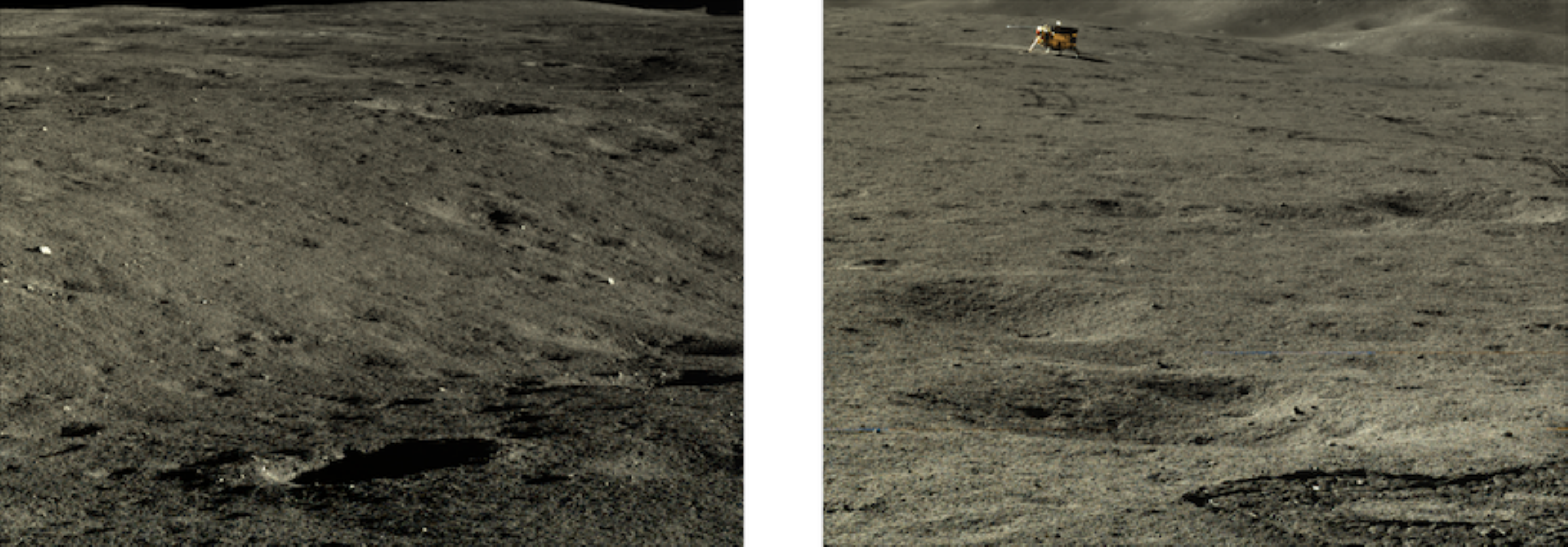}
   \caption{Images from the Moon. Panoramic camera of the rover. Chang'E-4.}
\label{fgr:moon_ce4}
\end{teaserfigure}

\ExplSyntaxOn
\newcommand\latinabbrev[1]{
  \peek_meaning:NTF . {
    #1\@}%
  { \peek_catcode:NTF a {
      #1.\@ }%
    {#1.\@}}}
\ExplSyntaxOff

\def\eg{e.g. }
\def\etal{et al. }


\begin{abstract}
Generative Adversarial Networks (GANs) \cite{Goodfellow14,Radford16} have had tremendous applications in Computer Vision. Yet, in the context of space science and planetary exploration the door is open for major advances. We introduce tools to handle planetary data from the mission Chang'E-4\footnote{Data and more information will be available at \href{https://www.github.com/decurtoidiaz/ce4/}{github.com/decurtoidiaz/ce4/}} and present a framework for Neural Style Transfer using Cycle-consistency \cite{Zhu17} from rendered images. The experiments are conducted in the context of the Iris Lunar Rover, a nano-rover that will be deployed in lunar terrain in 2021 as the flagship of Carnegie Mellon, being the first unmanned rover of America to be on the Moon.
\end{abstract}

\maketitle
\fancyfoot{}
\thispagestyle{empty}


\section{Introduction}
Generative Adversarial Network (GAN) \cite{Goodfellow14} are able to produce good quality high-resolution samples from images, both in the unscontrained and conditional setting \cite{Wu16,Yang17,Zhu17,Bousmalis17,Li17,Wang18,Wang18_2,Portenier18,Lombardi18,Yu18,Sankaranarayanan18,Romero18,Chen19}. Nonetheless, applications in the context of NASA missions and space exploration are scarce.
\\

Given the difficulty to handle planetary data we provide downloadable files in PNG format from the missions Chang'E-3 and Chang'E-4\footnote{Original PDS4 and PDS3 images and labels from missions to the Moon Chang'E can be obtained at \href{http://moon.bao.ac.cn/}{moon.bao.ac.cn}.}. In addition to a set of scripts to do the conversion given a different PDS4 Dataset. Example samples from the dataset can be seen in Figure \ref{fgr:moon_ce4}. We also provide the corresponding labels, where localization information is present. We run extensive experiments to train a model able to be used as a hyperrealistic feature of the current simulator used in the Iris Lunar Rover.

\label{sn:introduction}


\section{Overall System}
\label{sn:os}

Following the design principles and the perception pipeline proposed in \cite{Allan19} in the context of the NASA Mission Resource Prospector, we intend to design a simulator with hyperrealistic characteristics of the Moon that helps us deploy VIO/SLAM in a rover of the same characteristics. The intention is also that helps us address object detection and segmentation in this unmapped environment, where training data is very difficult and costly to obtain. Although at the present time data from the Moon is scarce, there are already some open datasets available in analogue environments such as the POLAR Stereo Dataset \cite{Wong17} that includes stereo pairs and LiDAR information or \cite{Vayugundla18}, that contains IMU, stereo pairs and odometry plus some additional localization data, all obtained on Mount Etna. Our intention is to provide downloadable files from the mission Chang'E-4 \cite{Zhang19} that could be easily used in CV and ML pipelines. We also provide scripts to handle alternate PDS4 Datasets. The context where this tools are being used is our specific sensor suite, that will be on-board the Iris Lunar Rover, a project led by Carnegie Mellon that intends to put forward a four pound rover into the surface of the Moon by 2021 and that will be America's first rover to explore the surface of the planet, consists on IMU, two high-fidelity cameras and odometry sensors. Furthermore, it also has a UWB module \cite{Ledergerber15,Mueller15,Alarifi16,Xu20} on-board to localize the rover with respect to the lander.


\section{Approach, long-term goal and prior work}
\label{sn:a}
Generative image generation is a key problem in Computer Vision and Computer Graphics. Variational Autoencoders (VAE) \cite{Kingma14,Lombardi18} try to solve the problem with an approach that builds on probabilistic graphical models. Autoregressive models (for instance PixelRNN \cite{Oord16}) have also achieved relative success generating synthetic images. In the past few years, Generative Adversarial Networks (GANs) \cite{Goodfellow14,Radford16,Odena17,Antoniou18,Wang16,Zhu16,Portenier18} have shown strong performance in image generation. Some works on the topic pinpoint the specific problem of scaling up to high-resolution samples \cite{Zhang17}, where conditional image generation is also studied while some recent techniques focus on stabilizing the training procedure \cite{Salimans16,Mescheder17,Mescheder18,Chen17,Dosovitskiy16,Zhao17,Karras18,Wei18,Brock19}. Other promising novel approaches include score matching with LANGEVIN sampling \cite{Song19,Song20} and the use of sequence transformers for image generation \cite{parmar18}.
\\

The use of these techniques though have seen little or no applications in space exploration and planetary research. We propose here a framework that could be used to generate abundant data of the Moon, Mars and other celestial bodies, so that learning algorithms could be trained on Earth and studied in simulation before being deployed in the real missions.
\\

The proposed approach consists on using a technique of Neural Style Transfer or Generative Image Generation, such as the criteria of cycle-consistency, together with an augmentation of the given dataset (in our case using data from the lunar missions Chang'E-3 and Chang'E-4, but the same applies to Mars or other planets) using GANs in the setting of unconstrained image generation.


\section{Cycle-consistent Generative Adversarial Networks}
\label{sn:ccn}

Our focus here is on Cycle-consistent Generative Adversarial Networks \cite{Zhu17}, where we work on unpaired image-to-image translation \cite{Park20}.
\\

Image-to-image translation is a type of problem in Computer Vision and Computer Graphics where the objective is to learn a correspondence function between an input sample and an output sample, using a training set of aligned or non-aligned image pairs.\\

More precisely, our goal is to learn a function
\begin{align}
G: X \to Y,
\end{align}
in a way that the distribution of samples $G(X)$ is as close as possible to the distribution $Y$. To accomplish this we are going to use and adversarial loss. Therefore, we couple it with the inverse correspondence
\begin{align}
F: Y \to X,
\end{align}
and use a criteria of cycle-consistency to address the fact that the problem is highly under constrained
\begin{align}
F(G(X)) \approx X \qquad and \qquad G(F(Y))\approx Y.
\end{align}
When we talk about paired training data, we refer to the fact that the training data consists of training examples $\{x_{c},y_{c}\}_{c=1}^{N}$, where the correspondences between $x_{c}$ and $y_{c}$ are given. Instead, we say that we are using unpaired training data, when the set consists of two training sets $\{x_{c}\}_{c=1}^{N}$ and $\{y_{a}\}_{a=1}^{N}$, where there is not explicitly given a correspondence between which $x_{c}$ corresponds to which $y_{a}$.
\\

Formally, the GAN objective \cite{Goodfellow14} involves finding a NASH equilibrium to the following two-player game:

\begin{align}
\min_{G} \max_{D} V(D,G) = \mathbb{E}_{x \sim p_{data}} \left[ \log D(x) \right] +\\
+ \mathbb{E}_{z \sim p_{z}} \left[ \log(1 - D(G(z)))\right],
\end{align}

\noindent
where $x$ is a ground truth image sampled from the true distribution $p_{data}$, and $z$ is a noise vector sampled from $p_{z}$ (that is, uniform or normal distribution). $G$ and $D$ are parametric functions where $G: p_{z} \rightarrow p_{data}$ maps samples from noise distribution $p_{z}$ to data distribution $p_{data}$. 
\\


\section{Neural Style Transfer}
\label{sn:nst}
Neural Style Transfer \cite{Gatys16,Gatys16_2,Ulyanov16} is based on the idea of synthesizing an original image by combining the content of one image together with the style of another sample. Here we will use cycle-consistent networks to attack this specific problem, with the aim of using a more general method that could help us solve concomitantly other tasks in the future. Moreover, the criteria of cycle-consistency assumes there is a bijection between the two domains, a constrain that could be often too restrictive, but that is very appropriate in our particular problem at-hand.


\section{Unconstrained Image Generation}
\label{sn:uig}
To tackle the problems that arise when training Cycle-consistent networks with a dataset with few samples, i.e. mainly mode collapse and artifacts, we propose to use Unconstrained Image Generation using GANs to enlarge the original dataset with unseen examples, that is, as a way to generate additional training samples that will help the learning procedure converge to the desired solution. To achieve this we make use of the construction developed in \cite{Brock19}.
\\


\section{Experiments}
\label{sn:ets}
\begin{figure}[H]
\centering
   \includegraphics[scale=0.2]{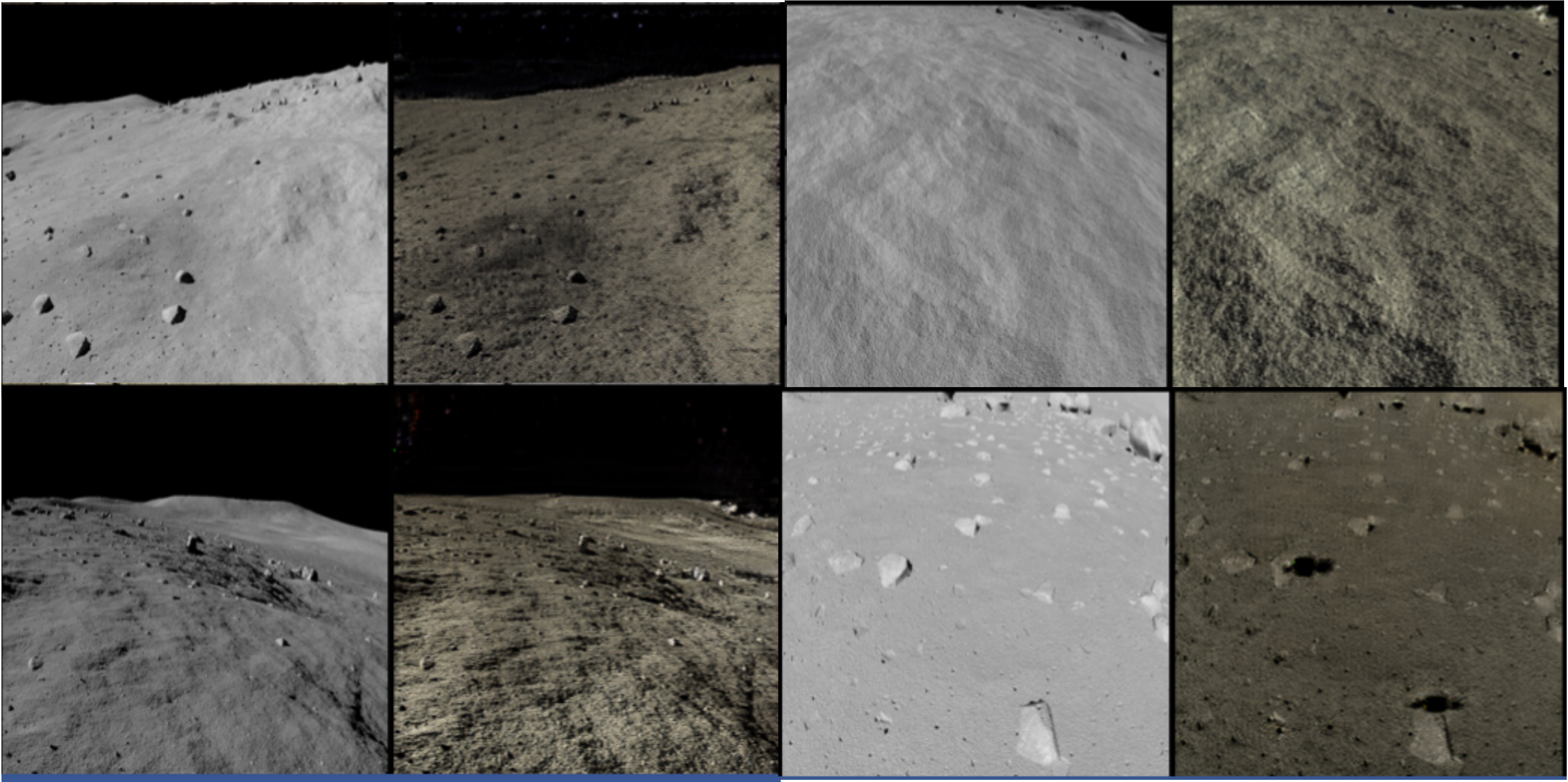}
   \caption{\textbf{Cycle-consistent gan}. Left: images from Kaggle, rendered simulator of the Moon. Right: style-Moon using our model. Trained at image size 256.}
\label{fgr:neural_moon_01}
\end{figure}

\begin{figure*}
\centering
   \includegraphics[scale=0.37]{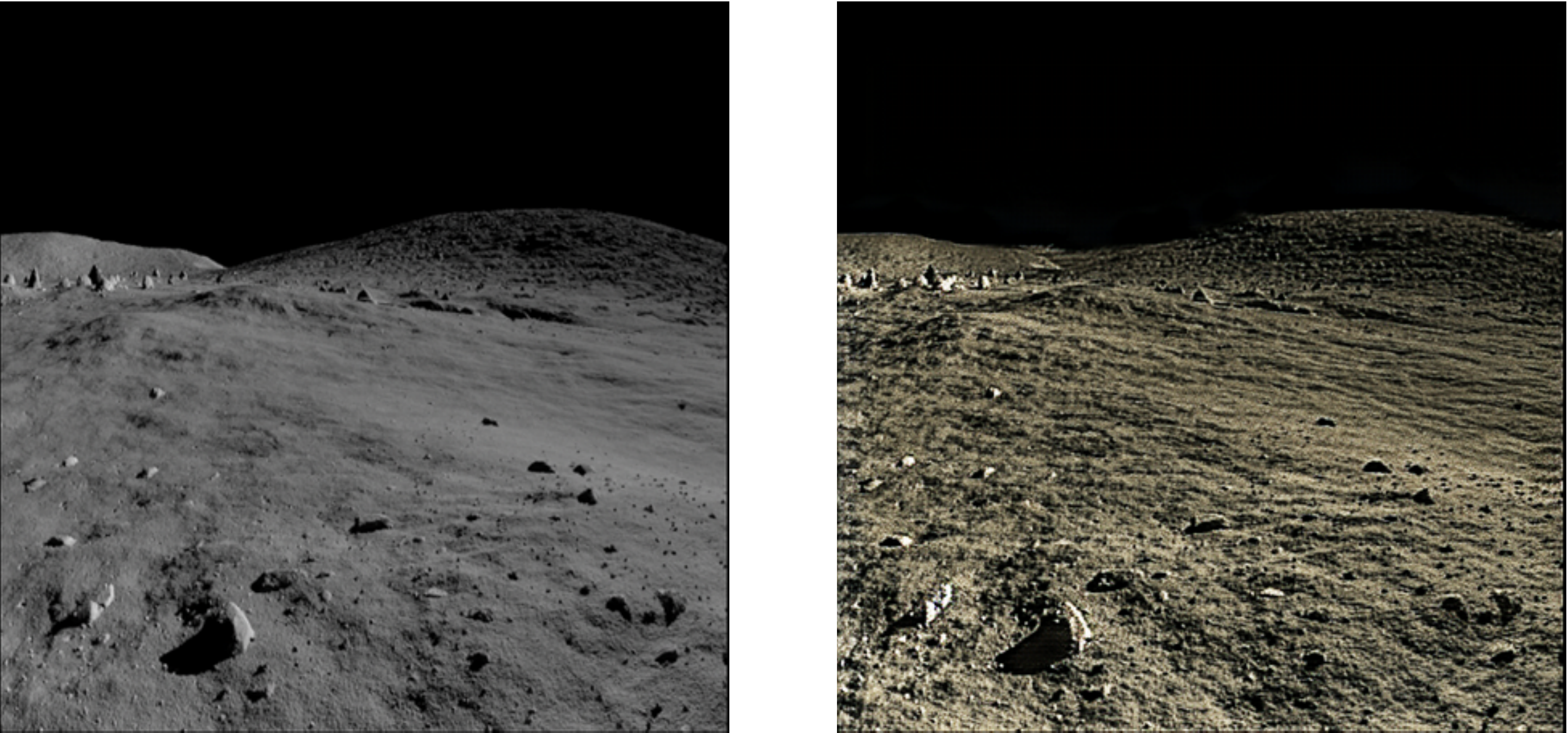}
   \caption{\textbf{Cycle-consistent gan}. Left: image from Kaggle, rendered simulator of the Moon. Right: style-Moon using our model. Trained at image size 512.}
\label{fgr:neural_moon_02}
\end{figure*}

Extensive experiments using data from Chang'E-3 and Chang'E-4 have been conducted, in particular we are using images from the panoramic camera of the rover and from the terrain camera of the lander. Some examples can be seen in Figures \ref{fgr:neural_moon_01} and \ref{fgr:neural_moon_02}, model trained at image size 256 and 512, respectively. As a source domain we are using samples from a rendered simulator of the Moon provided by Kaggle. The intention is to use the model in our actual renderer environment of the mission.

\section{Simulator}
\label{sn:s}
\begin{figure*}
\centering
   \includegraphics[scale=0.22]{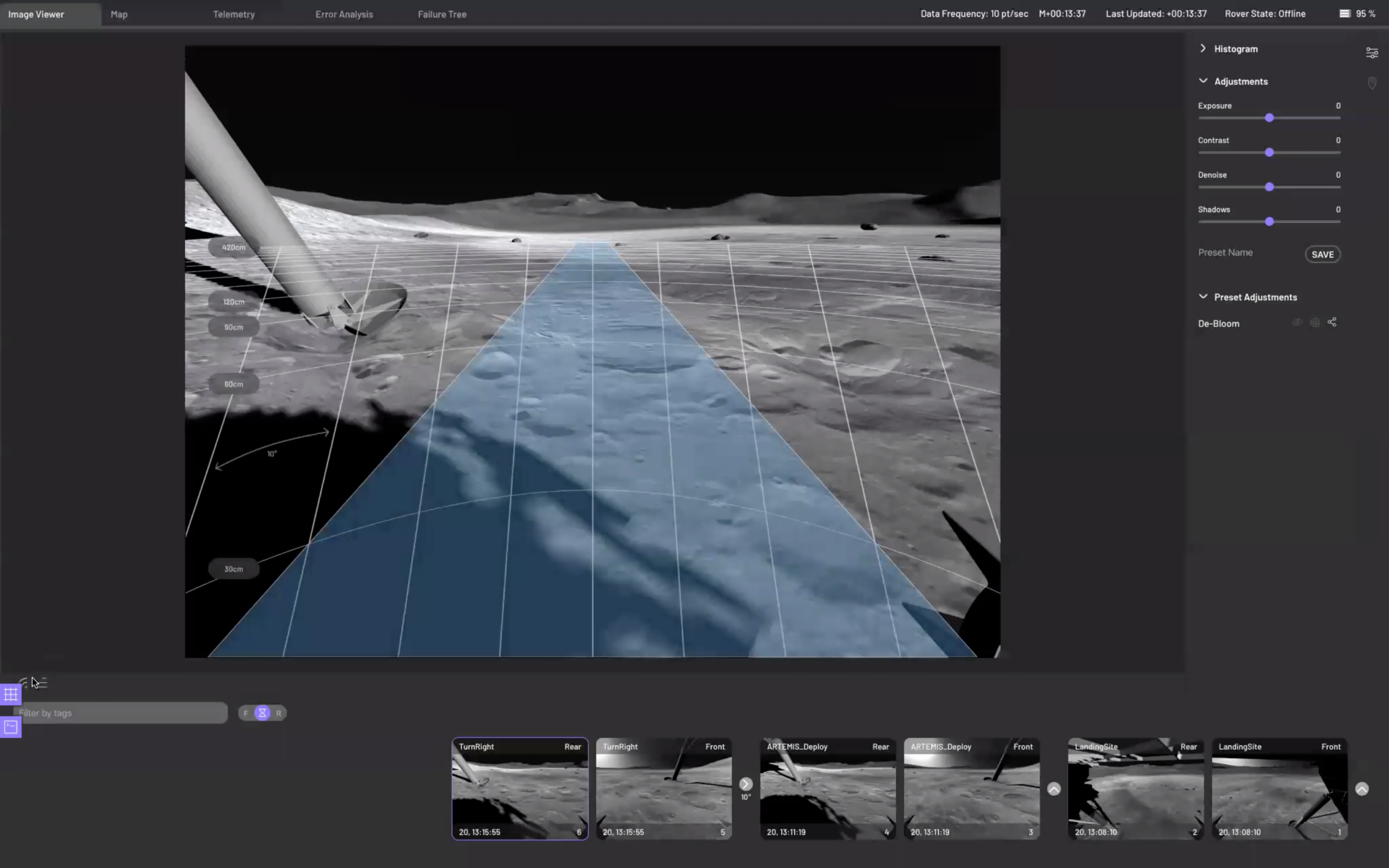}
   \caption{\textbf{Iris Lunar Rover}. Simulator used in the actual mission.}
\label{fgr:simulator_rover}
\end{figure*}

The context where this feature is being integrated is the actual simulator, see Figure \ref{fgr:simulator_rover}, of the Iris Lunar Rover, the rover of Carnegie Mellon that will fly to the Moon onboard the Peregrine Lander of Astrobotic in 2021. Data from the simulator will be of the utmost importance to train and test localization algorithms such as SLAM/VIO \cite{Schneider18,Usenko19}. The ability to have ample data to train will also amplify the capabilities of the modules designed for segmentation \cite{Badrinarayanan15,Chen17,Chen17_2,He17} and object detection \cite{Bolme10,Girshick15,Redmon16}. As well as to test the software design before the real mission.
 
\bibliographystyle{ACM-Reference-Format}

\end{document}